\documentclass[letterpaper, 10 pt, conference]{IEEEtran}  %

\IEEEoverridecommandlockouts
\usepackage{caption}
\usepackage{subcaption}
\usepackage{booktabs}
\usepackage{balance}
\usepackage{color}
\usepackage{tabularx}
\usepackage{graphicx} 
\usepackage{amsmath}
\usepackage{amssymb}
\usepackage{listings}
\usepackage{textcomp}
\usepackage{url}
\usepackage{multirow}
\usepackage{todonotes}
\usepackage{hyperref}
\usepackage{cleveref}
\usepackage{wrapfig}
\crefformat{footnote}{#2\footnotemark[#1]#3}
\usepackage{ragged2e} 
\usepackage[subnum]{cases}
\newcolumntype{Y}{>{\RaggedRight\arraybackslash}X} 
\usepackage{algorithm}
\usepackage[noend]{algpseudocode}
\usepackage{float}
\usepackage[bottom]{footmisc}
\usepackage{multicol, blindtext}
\usepackage{threeparttable}

\newcommand{\degree}{$^{\circ}$}

\title{\LARGE \bf
   Deploying COTS Legged Robot Platforms into a Heterogeneous Robot Team
}

\author{Benjamin Tam, Thomas Molnar, Fletcher Talbot, Brett Wood, Ryan Steindl%

\thanks{The authors are with the Robotics and Autonomous Systems Group, CSIRO, Pullenvale, QLD 4069, Australia. All correspondence should be addressed to {\tt\small benjamin.tam@csiro.au}}
}

\begin{document}
\maketitle
\thispagestyle{empty}
\pagestyle{empty}

\begin{abstract}
The recent availability of commercial-off-the-shelf (COTS) legged robot platforms have opened up new opportunities in deploying legged systems into different scenarios. While the main advantage of legged robots is their ability to traverse unstructured terrain, there are still large gaps between what robot platforms can achieve and their animal counterparts. Therefore, when deploying as part of a heterogeneous robot team of different platforms, it is beneficial to understand the different scenarios where a legged platform would perform better than a wheeled, tracked or aerial platform. Two COTS quadruped robots, Ghost Robotics' Vision 60 and Boston Dynamics' Spot, were deployed into a heterogeneous team. A description of some of the challenges faced while integrating the platforms, as well as some experiments in traversing different terrains are provided to give insight into the real-world deployment of legged robots.
\end{abstract}

\section{Introduction}
\label{sec:introduction}
The DARPA Subterranean (SubT) Challenge \cite{darpasubt} has highlighted current challenges in deploying autonomous robots into unknown environments. Due to severely limited communications, the robots are required to autonomously navigate terrain with little to no human input. Therefore, the robot platforms need to be stable and robust when faced with terrains from human-made urban environments to natural caves.
CSIRO Data61 is one of the teams competing in the challenge and has developed a heterogeneous team of robot platforms which use a common sensing payload, the \textit{CatPack} \cite{hudson2021heterogeneous}.
This enables different platforms to be integrated into the team and be deployed to different environments according to their strengths. To focus on multi-agent autonomy and the construction of a smaller tracked platform, the development of a legged robot platform Bruce \cite{steindl_2020} was halted in favour of acquiring commercial-off-the-shelf (COTS) legged platforms, such as the Ghost Robotics' Vision 60 \cite{ghostrobotics}, Boston Dynamics' Spot \cite{bostondynamics}, Unitree's Laikago \cite{unitree} or Anybotics' ANYmal C \cite{anymalc}. 

\section{Integration Challenges}
\label{sec:challenges}
The CatPack provides SLAM (simultaneous localisation and mapping) capability to the platforms using a lidar and IMU, and has cameras for object detection.
Point cloud and odometry data is then passed to the autonomy payload where a voxel occupancy map is created for global and local navigation. Either velocity commands or waypoints are sent via Robot Operating System (ROS) \cite{quigley_2009} to the robot platforms, allowing modularity for integration.
Both legged platforms deployed, Ghost Robotics' Vision 60 v4 \cite{ghostrobotics} and Boston Dynamics' Spot \cite{bostondynamics}, shown in Fig.~\ref{fig:robots}, are either ROS-enabled or have a ROS wrapper available for quick integration. Apart from the expected software integration and configuration parameter tuning, two challenges arose in deploying these platforms: 

\begin{figure}
\centering
\begin{subfigure}{.45\columnwidth}
  \centering
    \includegraphics[width=0.9 \columnwidth]{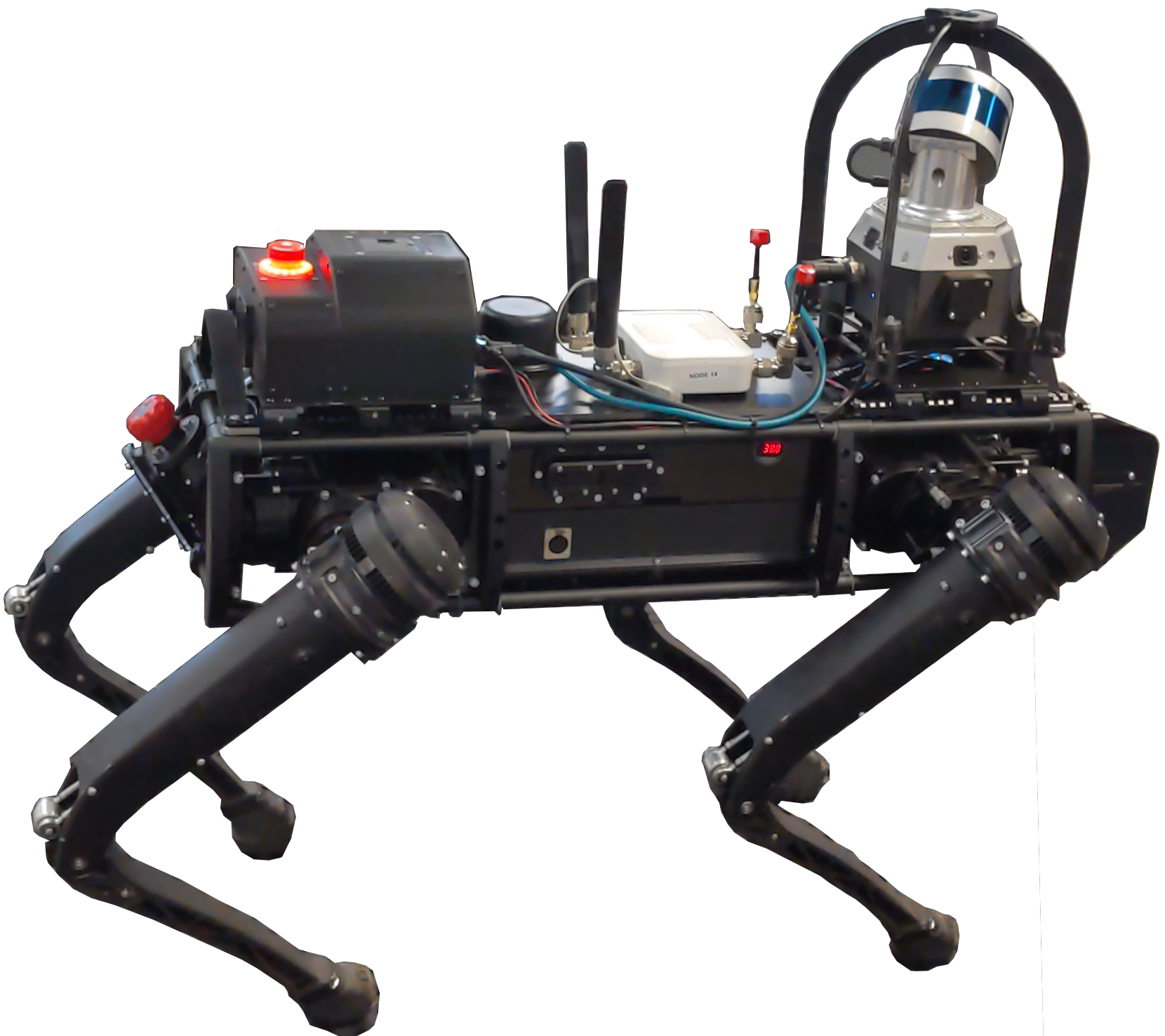}
    \caption{}
    \label{fig:ghost}
\end{subfigure}%
\begin{subfigure}{.45\columnwidth}
  \centering
    \includegraphics[width=1.05 \columnwidth]{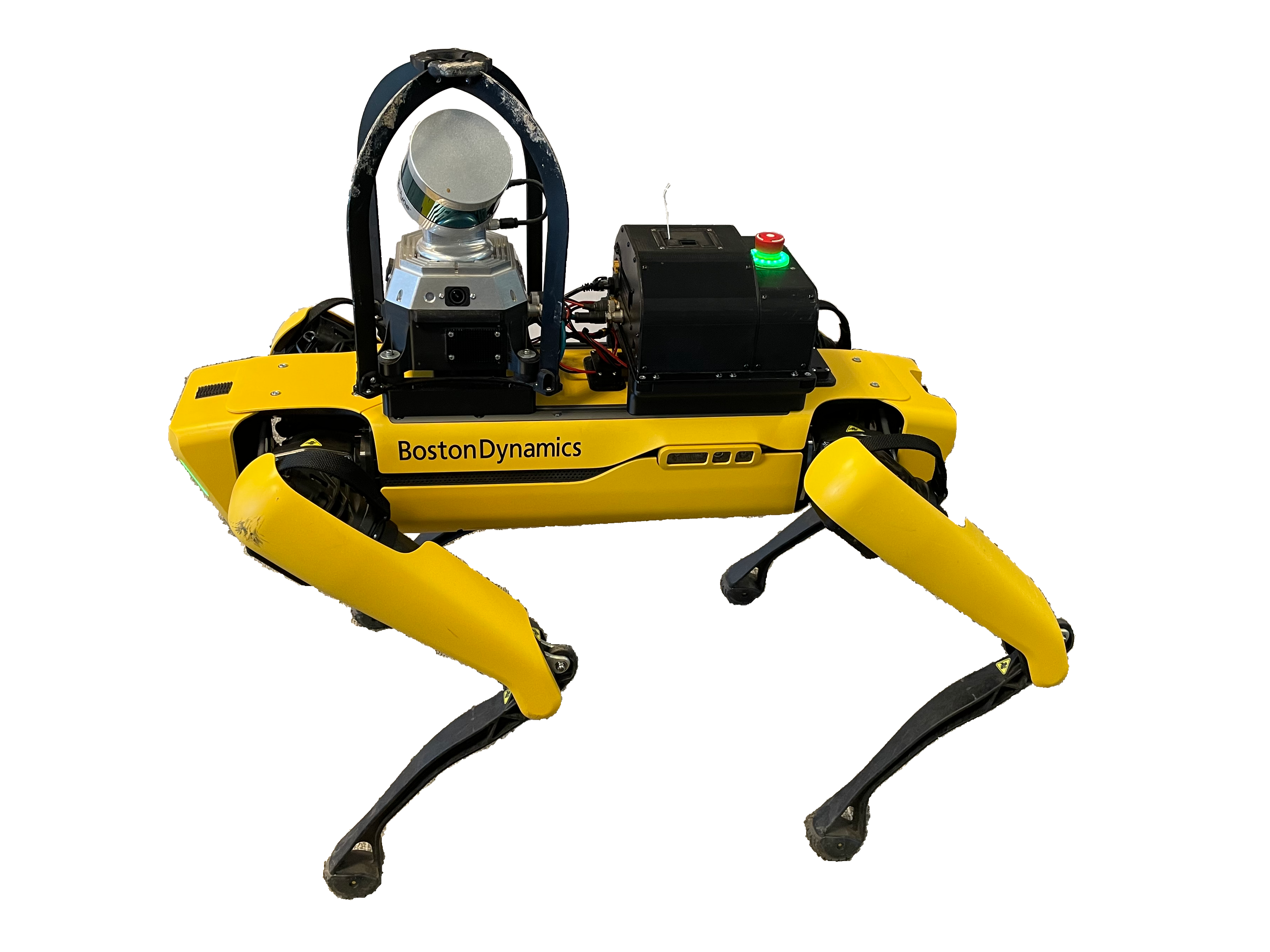}
    \caption{}
    \label{fig:ghost}
\end{subfigure}%
\caption{a) Ghost Robotics Vision60 v4.0, and b) Boston Dynamics Spot quadruped platform.}
\label{fig:robots}
\end{figure}

\begin{figure*}
\centering

\begin{subfigure}{.24\textwidth}
  \centering
  \includegraphics[width=.99\linewidth]{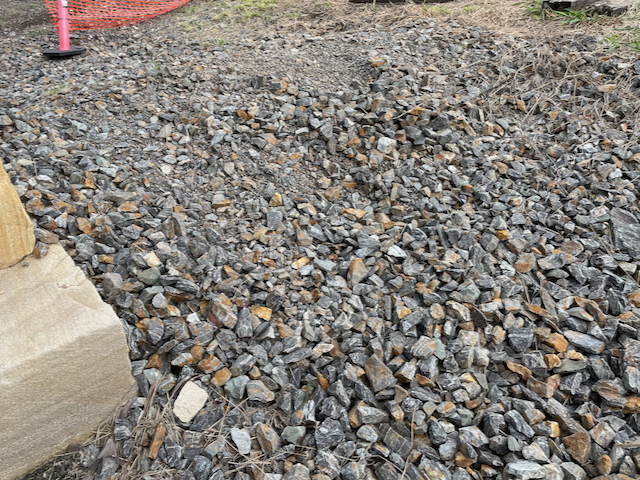}
   \caption{}
  \label{fig:rockA}
\end{subfigure}%
\begin{subfigure}{.24\textwidth}
  \centering
  \includegraphics[width=.99\linewidth]{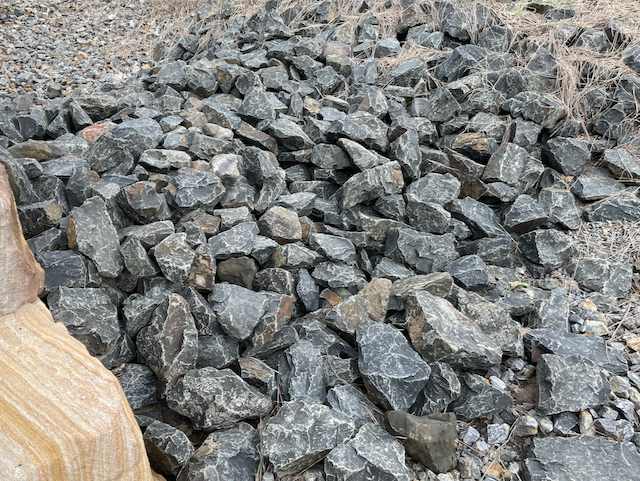}
  \caption{}
  \label{fig:rockB}
\end{subfigure}%
\begin{subfigure}{.24\textwidth}
  \centering
  \includegraphics[width=.99\linewidth]{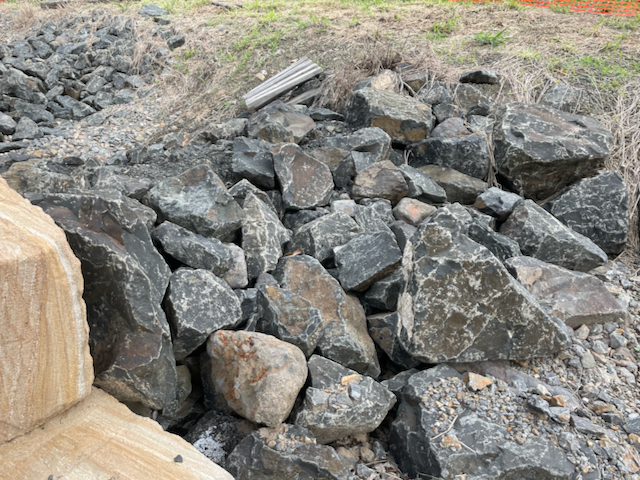}
  \caption{}
  \label{fig:rockC}
\end{subfigure}
\begin{subfigure}{.24\textwidth}
  \centering
  \includegraphics[width=0.99 \linewidth]{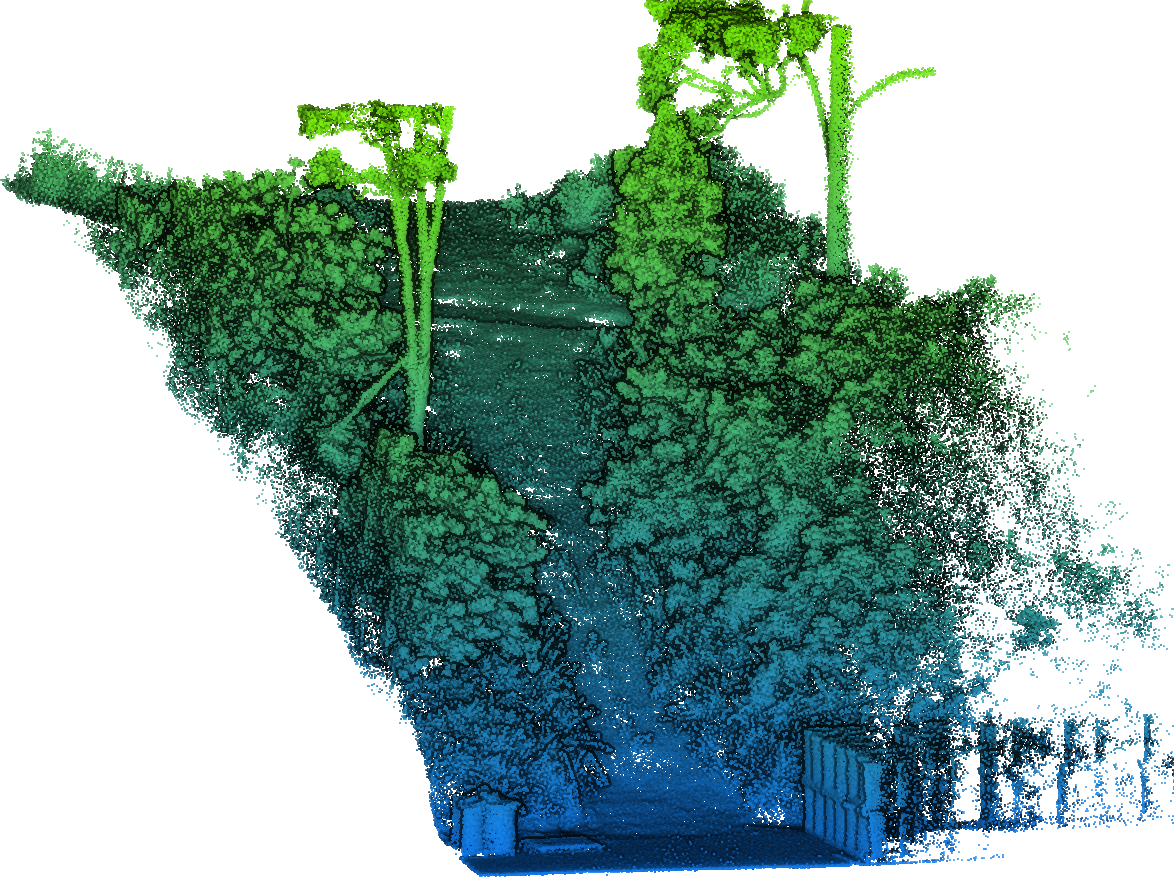}
  \caption{}
  \label{fig:dirtrd}
\end{subfigure}%
\caption{Rock piles containing rocks of size: a) A - 0.01-0.1\,m; b) B - 0.1-0.3\,m; and c) C - 0.2-0.5\,m. d) Point cloud generated by the CatPack of the dirt and loose rock road with local slope of 15 - 25\degree.}
\vspace{-0.5cm}
\label{fig:rock_piles}
\end{figure*}

\textit{Communications} - Subterranean environments degrade wireless communications due to limited direct line of sight between robots. Therefore, bandwidth is prioritised for essential multi-agent map sharing and coordination data. The team uses a network of Rajant radios for communication.
As the legged platforms were designed for remote user operation via a mobile app, each robot has their own wireless access point. These additional wireless systems working in the same band (2.4G\,Hz and 5\,GHz) as the Rajant radios caused interference. Thus, these controllers were switched off during missions to reduce radio noise. Additional issues on the Vision 60 platform was local robot control network packets being accidentally sent via the Rajant radios and other robot platforms acquiring an IP address from the DHCP server onboard, causing further network bandwidth decrease. Unwanted packets were resolved via using a DMZ, while the onboard DHCP server was disabled and static IP addresses assigned.

\textit{Payload Mounting} -
The payload mounting rails for the Vision 60 are at the front and rear of the robot, while Spot has them at the middle. Due to the 45\degree (off the horizontal) mounting of the lidar on the CatPack, the mounting on the Spot near the middle occludes the view of the area in front of the robot, compared to the Vision 60 where the CatPack could be mounted near the front. 
Thus, Spot's onboard depth sensors where used to fill in this missing information.
\section{Experiments}
\label{sec:experiments}

Previous works in testing COTS legged robots capability \cite{adam_spot_2020} have shown that while they are capable of traversing challenging unstructured terrain, their reliability on these terrain is still limited. Thus, experiments were conducted to identify terrains legged robot platforms can reliably be deployed to.
A set of challenging terrains were traversed to test the capabilities of the legged platforms compared to the tracked platform BIA5's ATR. Note that while it is not a like-for-like comparison of legged versus tracked platform, the experiments provide real world scenarios for the deployment of such platforms. Table~\ref{tab:platform_capabilities} compares the specifications of the platforms. Also, the Vision 60 has onboard vision sensors but were not used due to the code being experimental, while Spot has a more mature vision system.
The different types of terrain traversed include on three different sized rock piles (Fig.~\ref{fig:rockA}-\ref{fig:rockC}), a sloped road (Fig.~\ref{fig:dirtrd}), doorways and stairs. 

\begin{table}[b]
\vspace{-0.3cm}
\caption{Platform Capabilities}
\label{tab:platform_capabilities}
\centering
\vspace{-0.1cm}
\begin{tabular}{@{}lccc}
\toprule

 \multirow{2}{*}{Attribute}  & GR & Boston Dynamics & BIA5 \\ 
    & Vision 60 & Spot & ATR  \\ 

\midrule
Actuators & 12 & 12 & 2 \\
Weight with payload (kg) & 48.4 & 38.1 & 90  \\
Width (m)  & 0.55 & 0.50 & 0.78  \\
Length (m) & 0.93 & 1.1 & 1.4  \\
Max Step Height (m) & 0.25 & 0.3 & 0.25\\
Max Slope (\degree) & x & 30 & 43 \\
\bottomrule
\end{tabular}
\end{table}

\section{Discussion}
\label{sec:discussion}

For the sloped dirt road, the Vision 60 was unable to traverse it successfully, while both the Spot and ATR were able to. This could be due to Vision 60 being heavier and does blind mode locomotion, causing the legs to impact the ground with larger forces. This caused the loose dirt and rocks to give way under the rubber feet resulting in the robot slipping and falling over. This was not observed on the lighter Spot or the large contact footprint area of the tracks on the ATR. For the rock piles, the Vision 60's blind mode locomotion was able to reliably traverse rock pile A, while it had mixed results for rock pile B. This was not the case with Spot, where the perception system was able to place the feet on better footholds for both rock pile A and B. In comparison, the ATR was capable of traversing all three piles. It was also observed that the Vision 60 would naturally strafe down an incline when walking across the slope which required manual compensation. 

The legged robots were able to traverse narrow corridors and doorways easier than the tracked platform using the same local navigation algorithm, but with a footprint reflecting the smaller robot size. As the doors had a minimum 0.8\,m width, the tolerance for the ATR was smaller than the legged platforms, causing the local navigation algorithm to fail to plan through.

Both legged platforms have a dedicated stairs mode which changes gait parameters for going up stairs forwards and down stairs backwards with good reliability. The ATR can traverse up or down a single flight of stairs, but is unstable and could topple over if the velocity commands are not precise. For multiple flights of stairs, the ATR's footprint causes difficulties in turning around on the landing, something the legged platforms do not suffer from. 

\section{Conclusions}
\label{sec:conclusions}
Through testing the legged platforms on various terrains, it was found that their strength is in being deployed onto stairs and narrow passageways. While the legged platforms were able to traverse unstructured outdoor terrain, the tracked robot provided a more reliable platform for traversing these terrains where the ground surface could be loose.

\section*{Acknowledgements}
The authors would like to thank the CSIRO Data61 SubT Challenge Team.

\balance

\bibliographystyle{IEEEtran}
\bibliography{references}

\end{document}